\documentclass[8pt]{article}
\usepackage[margin=20mm]{geometry}

\usepackage{cite}
\usepackage{amsmath,amssymb,amsfonts}
\usepackage[ruled]{algorithm2e}
\usepackage{graphicx}
\usepackage{textcomp}
\usepackage{xcolor}
\usepackage{url}
\usepackage{hyperref}
\usepackage{comment}
\usepackage{booktabs}
\usepackage{tabularx}
\def\BibTeX{{\rm B\kern-.05em{\sc i\kern-.025em b}\kern-.08em
　　T\kern-.1667em\lower.7ex\hbox{E}\kern-.125emX}}
\begin{document}

\title{Enhancing In-vehicle Multiple Object Tracking Systems with Embeddable Ising Machines}

\author{
Kosuke Tatsumura$^{1,\dag,\ast}$, Yohei Hamakawa$^{1,\dag}$, Masaya Yamasaki$^{1}$, \\
Koji Oya$^{2}$, and Hiroshi Fujimoto$^{2}$ \\
\small $^{1}$Corporate Research and Development Center, Toshiba Corporation, Kawasaki 212-8582, Japan\\
\small $^{2}$MIRISE Technologies, Aichi 470-0111, Japan\\
\small $^{\dag}$These authors contributed equally to this work.\\
\small $^{\ast}$Corresponding author: Kosuke Tatsumura (e-mail: kosuke.tatsumura@toshiba.co.jp)
}
\date{}

\maketitle

\begin{abstract}
A cognitive function of tracking multiple objects, needed in autonomous mobile vehicles, comprises object detection and their temporal association. While great progress owing to machine learning has been recently seen for elaborating the similarity matrix between the objects that have been recognized and the objects detected in a current video frame, less for the assignment problem that finally determines the temporal association, which is a combinatorial optimization problem. Here we show an in-vehicle multiple object tracking system with a flexible assignment function for tracking through multiple long-term occlusion events. To solve the flexible assignment problem formulated as a nondeterministic polynomial time-hard problem, the system relies on an embeddable Ising machine based on a quantum-inspired algorithm called simulated bifurcation. Using a vehicle-mountable computing platform, we demonstrate a realtime system-wide throughput (23 frames per second on average) with the enhanced functionality. 
\end{abstract}

\section{Introduction}\label{sec:introduction} 

Autonomous control in mobile vehicles for advanced driver assistance systems (ADAS) or toward fully automated driving is realized with high-speed realtime systems that periodically execute a task pipeline comprising sensing, understanding, planning, and controlling~\cite{Yurtsever20, Levinson11, Nakade23}. Understanding circumstances and ego motion is advanced information processing including object detection and tracking \cite{SORT,DeepSORT,JDE,FiarMOT,GIAOTracker,UniTrack,ByteTrack,TransMOT}, localization and mapping~\cite{Mur-Artal17}, etc. Those information processing should be achieved efficiently with vehicle-mountable computing platforms~\cite{Su21, Yamada20, Fujii18} under constraints of size, power, and cost.

\begin{figure}[t]
\centering
\includegraphics[width=17.2 cm]{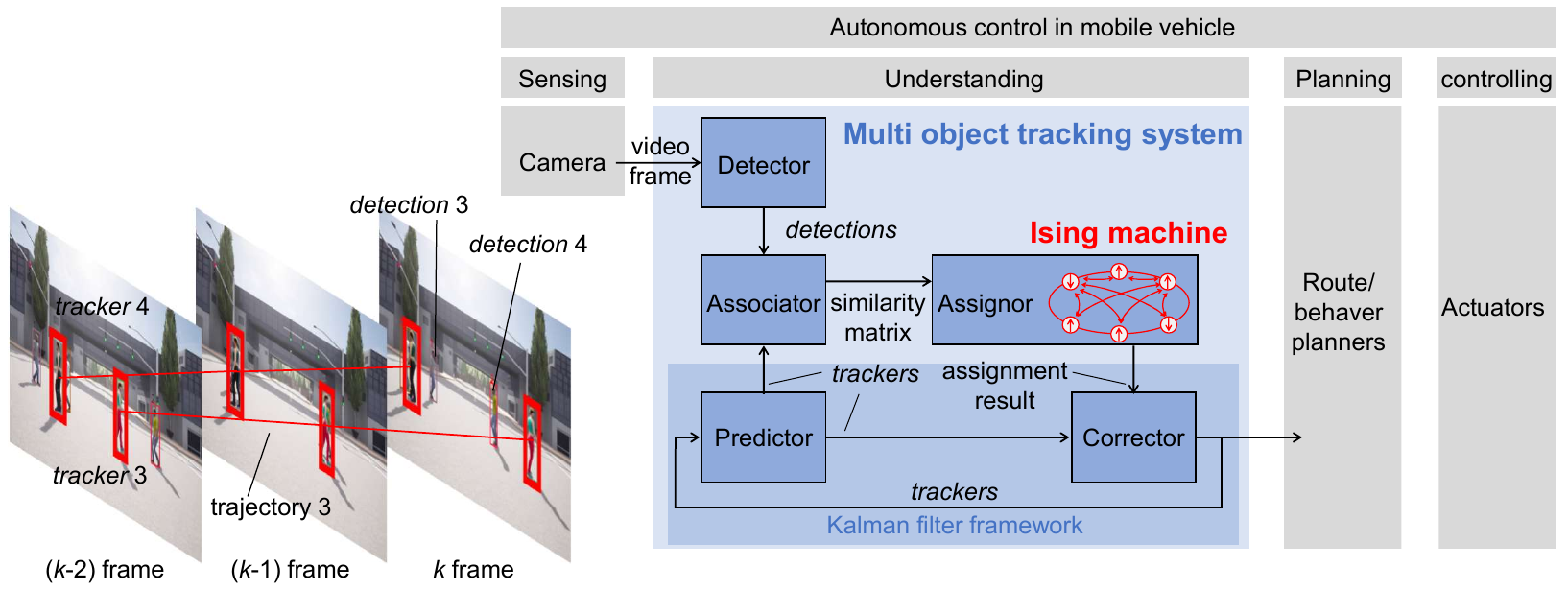}
\caption{In-vehicle multiple object tracking (MOT) system using an embeddable Ising machine to solve a flexible assignment problem formulated as NP-hard combinatorial optimization.}
\label{Fig_MOTsystem}
\end{figure}

Multiple object tracking (MOT) is a cognitive process that identifies and tracks multiple objects despite movement of the objects, even through temporary occlusion events such as crossing of the objects. Modern MOT systems~\cite{SORT,DeepSORT,JDE,FiarMOT,GIAOTracker,UniTrack,ByteTrack,TransMOT} use tracking-by-detection strategy, where the objects that have been identified and chased by the system (hereafter, \textit{trackers}) are associated with the objects detected at a current video frame (hereafter, \textit{detections}) and then updated using the information of the matched \textit{detections} (also see Fig.~\ref{Fig_MOTsystem}). The \textit{trackers}, thus, depict the trajectories of the objects. The assignment (matching) between \textit{trackers} and \textit{detections} is determined based on a similarity (or association) matrix, where each matrix element corresponds to the similarity between a \textit{tracker} and a \textit{detection}. Various sophisticated definitions of similarity have been recently proposed using advanced machine learning methodology~\cite{SORT,DeepSORT,JDE,FiarMOT,GIAOTracker,UniTrack,ByteTrack,TransMOT}. The assignment problem to maximize the overall likelihood, which finally determines the temporal association of objects, is a combinatorial optimization problem, more specifically a bipartite graph matching problem. The assignment problems in those MOT systems~\cite{SORT,DeepSORT,JDE,FiarMOT,GIAOTracker,UniTrack,ByteTrack,TransMOT} that assume one-to-one correspondence between \textit{trackers} and \textit{detections} are linear and thus can be solved in polynomial times by an exact algorithm called the Hungarian algorithm~\cite{Kuhn55}. When considering the period of object crossing (occlusion), many-to-one correspondence (many \textit{trackers} to one \textit{detection)} could be more feasible. However, those assignment problems considering the possibility of many-to-one correspondence as well are difficult combinatorial optimization problems.

Ising machines are domain-specific computers that aim at solving difficult combinatorial optimization problems in short time~\cite{Finocchio19, sbm1, FPL19, sbm2, NatEle, kanao23,Kashimata24,Matsumoto22, johnson11,king23, honjo21, kalinin20, PoorCIM19, cai20, borders19, aadit22, litvinenko23, Graber24, moy22, albertsson21, wang21, sharma22, kawamura23, matsubara20, waidyasooriya21, okuyama19,SimCIM21}. The Ising machines seek the lowest energy-states of Ising spin models~\cite{brush67} that consist of binary variables, called spins, coupled each other with pairwise interactions. The Ising problem, equivalent to quadratic unconstrained binary optimization (QUBO), belongs to the nondeterministic polynomial time (NP)-hard class~\cite{barahona82, lucas14}; various types of computationally-hard combinatorial problems can be formulated as the Ising problem~\cite{lucas14}. The Ising machines have been implemented with a variety of hardware~\cite{Finocchio19} including superconducting qubits~\cite{johnson11,king23}, optical systems~\cite{honjo21, kalinin20, PoorCIM19}, memristor-based neural networks~\cite{cai20}, probabilistic bits~\cite{borders19,aadit22}, spintronics systems~\cite{litvinenko23}, coupled oscillators~\cite{Graber24,moy22,albertsson21,wang21,sbm1}, analog computing units~\cite{sharma22}, application specific integrated circuits (ASICs)~\cite{kawamura23,matsubara20}, field programmable gate arrays (FPGAs)~\cite{sbm1, FPL19, sbm2, NatEle, Kashimata24, Matsumoto22, waidyasooriya21,SimCIM21}, and graphics processing units (GPUs)~\cite{sbm1, sbm2, okuyama19}.

In-vehicle computing platforms for autonomous control~\cite{Su21, Yamada20, Fujii18} need to be equipped with parallel and programmable coprocessors like embedded FPGAs/GPUs/neural processing unit (NPUs) to efficiently execute various and computationally-intensive workloads. Of various Ising machines, a part of ones~\cite{sbm1, FPL19, sbm2, NatEle, waidyasooriya21, SimCIM21, okuyama19} are based on highly-parallelizable algorithms (not limited to special hardware) and thus can be implemented/accelerated with in-vehicle parallel coprocessors. Those embeddable Ising machines may enable more rational/more functional information processing based on NP-hard combinatorial optimizations for ADAS/automated driving. Some literature~\cite{Govaers21,Zaech22,McCormick22} has investigated the applicability of quantum mechanics-based Ising machines (quantum annealers) to the assignment problems in MOT assuming one-to-one correspondence. Other literature has reported centralized (out-vehicle) systems using quantum annealers for traffic flow optimization~\cite{Neukart17} or swarm robot control~\cite{Ohzeki19}. High-speed financial trading systems with embeddable Ising machines~\cite{ISCAS20,ACCESS23a,ACCESS23b} have been also demonstrated. However, in-vehicle systems with Ising machines have not been studied.

Here we propose and demonstrate an in-vehicle high-speed realtime system enhanced with an embeddable Ising machine, which executes a MOT procedure featuring a flexible assignment function based on NP-hard combinatorial optimization.

With the flexible assignment, the proposed system realizes object tracking through multiple long-term occlusion events. The assignment problem between \textit{detections} and \textit{trackers} is formulated as a QUBO problem, whose total cost function is a linear combination of an objective function to maximize the overall likelihood and a penalty function corresponding to the constraint for one-to-one correspondence. The system solves the QUBO problem twice per video frame while changing the weight coefficient for the penalty function and then detects the occurrences and their locations of occlusion events as the difference between the two solutions (i.e., the two assignment tables), where an assignment table with many \textit{trackers} being matched to one \textit{detection} (a constraint-violation solution) can be selected if it is more feasible in terms of the total cost function upon the execution with the smaller weight coefficient. 

The system uses an embeddable Ising machine based on a quantum-inspired algorithm called simulated bifurcation (SB)~\cite{sbm1, FPL19, sbm2, NatEle, kanao23,Kashimata24,Matsumoto22}. The algorithm of SB was derived in 2019~\cite{sbm1} through classicizing a quantum-mechanical Hamiltonian describing a quantum adiabatic optimization method~\cite{qbm} and improved in 2021~\cite{sbm2}, which numerically simulates the time-evolution of a classical nonlinear oscillator network exhibiting bifurcation phenomena, where two branches of the bifurcation in each oscillator correspond to two states of each Ising spin. The operational mechanism of SB is based on an adiabatic and ergodic search~\cite{sbm1}. The MOT system is implemented with two vehicle-mountable (middle-range) FPGAs; one is for object detection and the other for the assignment by SB-based Ising machine. We demonstrate a realtime system-wide throughput of more than 20 frames per second for the Ising machine-enhanced MOT functionality.

\section{Results}\label{sec_results}

\subsection{Flexible assignment}\label{sec_FxAssignment}

\begin{figure}[t]
\centering
\includegraphics[width=17.2 cm]{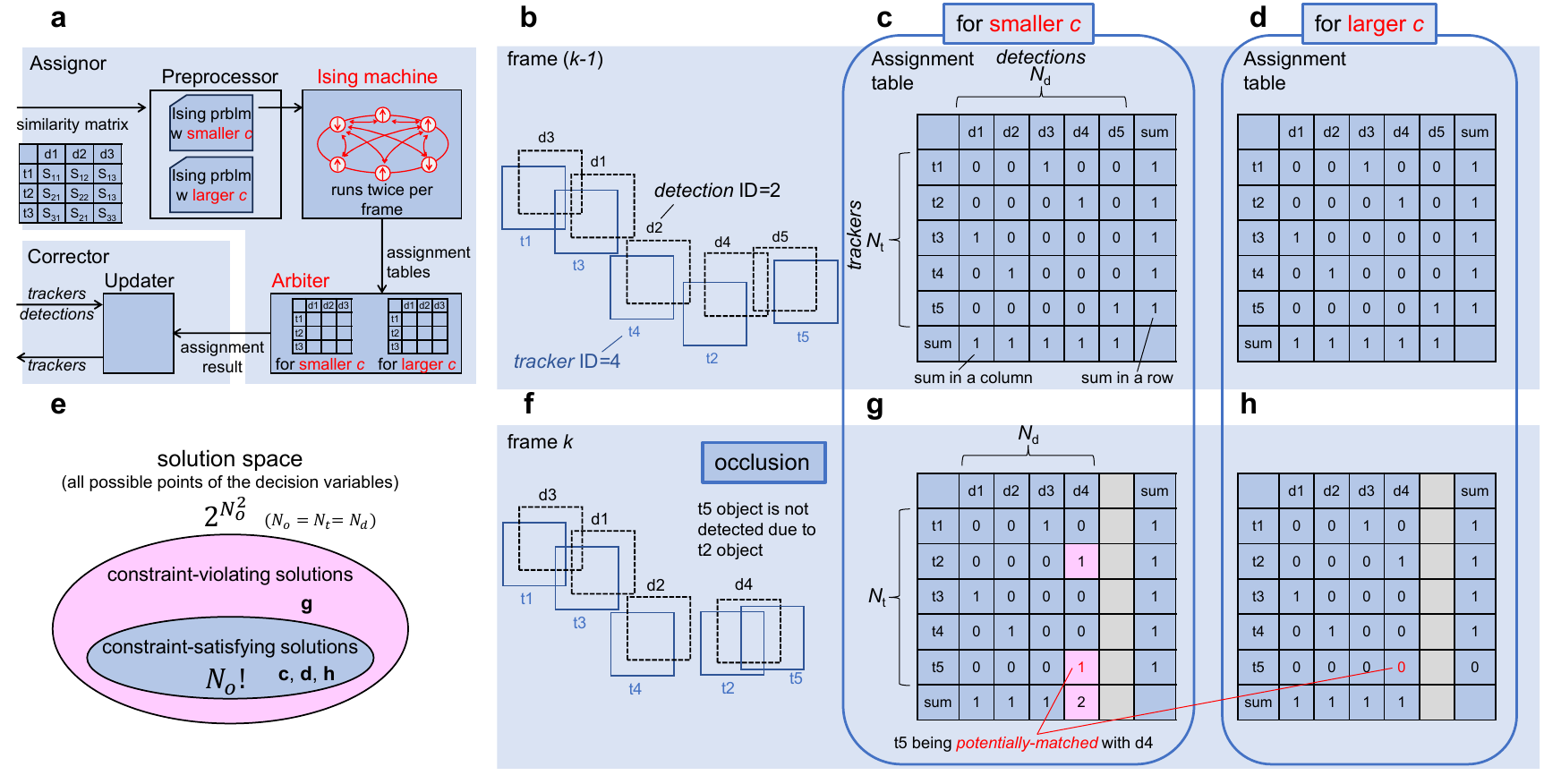}
\caption{Flexible assignment function in the MOT system. (\textbf{a}) Submodules in the assignor. The assignment result between \textit{trackers} and \textit{detections} considering the possibility of many-to-one correspondence is determined from two assignment tables that the Ising machine produces by solving the similarity matrix-based assignment problem two times per frame while changing the weight coefficient of the penalty function corresponding to the constraint for one-to-one correspondence. (\textbf{b}) A Scene without occlusion and the resultant assignment tables (\textbf{c} and \textbf{d}). (\textbf{f}) A scene with occlusion and the resultant assignment tables (\textbf{g} and \textbf{h}). (\textbf{e}) Solution space of the decision variables including constraint-violating/satisfying solutions.}
\label{Fig_FxMatching}
\end{figure}

A similarity matrix $S$ (also referred to as an association matrix in other literature) is defined as some kind of distance in real and/or feature spaces for all pairs, ($t$, $d$), between \textit{trackers} and \textit{detections} with each matrix element $S_{t,d}$ being a real number. Based on the similarity matrix, an assignment function under the constraint for one-to-one correspondence determines an assignment table with each element $b_{t,d}$ being a binary decision variable corresponding to $match$ or $unmatch$. The proposed flexible assignment function (Fig.~\ref{Fig_FxMatching}), using two binary assignment tables, introduces the third state of $potentially$-$match$ to represent a \textit{tracker} being matched with an occluded object (a hidden \textit{detection}). The state of $potentially$-$match$ is utilized to better estimate the dynamics of \textit{trackers} as discussed in the next section. 

We start with a QUBO formulation of the assignment problem assuming one-to-one correspondence. Given $N_{t}$ \textit{trackers} and $N_{d}$ \textit{detections} at a video frame, we prepare $N_{t}N_{d}$ binary variables ($b_{t,d}$) with each being defined as follows.

\begin{equation}\label{Eq_def_b}
b_{t,d}=
\begin{cases}
1, & (\text{if}\; t\text{th and } d\text{th objects are matched}),\\
0, & (\text{if}\; t\text{th and } d\text{th objects are unmatched}).
\end{cases}
\end{equation}

As a combinatorial optimization problem, we search for the bit configuration $\{b_{t,d}\}$ (the assignment table) that minimizes a total cost function $H_{\mathrm{cost}}$.
$H_{\mathrm{cost}}$ is a linear combination of an objective function $H_{\mathrm{object}}$ to maximize the overall likelihood and a penalty function, $H_{\mathrm{penalty1}}+H_{\mathrm{penalty2}}$, corresponding to the constraint for one-to-one correspondence:

\begin{equation} \label{Eq_Cost}
H_{\mathrm{cost}}=H_{\mathrm{object}}+c(H_{\mathrm{penalty1}}+H_{\mathrm{penalty2}}),
\end{equation}

\begin{equation} \label{Eq_Obj}
H_{\mathrm{object}}= - \sum_{d=1}^{N_d}\sum_{t=1}^{N_t}S_{t,d}b_{t,d},
\end{equation}

\begin{equation} \label{Eq_pnlty1}
H_{\mathrm{penalty1}}=
\begin{cases}
\sum\limits_{d=1}^{N_d} \Bigl( \sum\limits_{t=1}^{N_t} b_{t,d}-1 \Bigr) ^2 & (\text{if}\; {N_t} \ge {N_d}),\\
\sum\limits_{d=1}^{N_d} \Bigl( \sum\limits_{t\neq t^{\prime}} b_{t,d}b_{t^{\prime},d} \Bigr) & (\text{if}\; {N_t} < {N_d}),
\end{cases}
\end{equation}

\begin{equation} \label{Eq_pnlty2}
H_{\mathrm{penalty2}}=
\begin{cases}
\sum\limits_{t=1}^{N_t} \Bigl( \sum\limits_{d=1}^{N_d} b_{t,d}-1 \Bigr) ^2 & (\text{if}\; {N_t} \le {N_d}),\\
\sum\limits_{t=1}^{N_t} \Bigl( \sum\limits_{d\neq d^{\prime}} b_{t,d}b_{t,d^{\prime}} \Bigr) & (\text{if}\; {N_t} > {N_d}),
\end{cases}
\end{equation}
where $c$ in Eq.~\ref{Eq_Cost} is a weight coefficient of the penalty function to the objective function. The constraint for one-to-one correspondence is explicitly expressed as equality constraints:

\begin{equation} \label{Eq_const1}
\begin{cases}
\sum\limits_{t=1}^{N_t} b_{t,d}=1 \ \ \text{for all}\ d & (\text{if}\; {N_t} \ge {N_d}),\\
\sum\limits_{t\neq t^{\prime}} b_{t,d}b_{t^{\prime},d}=0 \ \ \text{for all}\ d & (\text{if}\; {N_t} < {N_d}),
\end{cases}
\end{equation}

\begin{equation} \label{Eq_const2}
\begin{cases}
\sum\limits_{d=1}^{N_d} b_{t,d}=1 \ \ \text{for all}\ t & (\text{if}\; {N_t} \le {N_d}),\\
\sum\limits_{d\neq d^{\prime}} b_{t,d}b_{t,d^{\prime}}=0 \ \ \text{for all}\ t & (\text{if}\; {N_t} > {N_d}).
\end{cases}
\end{equation}

The objective function $H_{\mathrm{object}}$ is minimized when the sum of $S_{t,d}$ for the pairs of matched \textit{trackers} and \textit{detections} (with $b_{t,d}$ being 1) is maximized. The penalty function, $H_{\mathrm{penalty1}}\!+\!H_{\mathrm{penalty2}}$, is minimized when satisfying the constraint for one-to-one correspondence. Constraint violations increase the penalty, with $(H_{\mathrm{penalty1}}\!+\!H_{\mathrm{penalty2}})\!=\!0$ if there are no violations. In the case of ${N_t} \!>\! {N_d}$, \textit{trackers} having no corresponding \textit{detections} (one-to-zero correspondence) are allowed. This happens, for example, when there are candidates of \textit{trackers} to be deleted (ex. frame-out objects). In the case of ${N_t} \!<\! {N_d}$, \textit{detections} having no corresponding \textit{trackers} (zero-to-one correspondence) are also allowed. This is the case, mostly, for the situation including candidates of additional \textit{trackers} (ex. frame-in objects). Any double coincidences (many-to-one correspondence) correspond to the constraint violations.

In the flexible assignment function that enables detecting the occurrences and their locations of occlusion events, as illustrated in Fig.~\ref{Fig_FxMatching}\textbf{a}, we solves the QUBO problem twice per video frame (per similarity matrix) with two weight coefficients, larger $c$ and smaller $c$, by the Ising machine, and then obtain two assignment tables. The table for the larger $c$ more likely satisfies the constraint for one-to-one correspondence, while one for the smaller $c$ more likely tolerates many-to-one assignments (constraint violations). We arbitrage the two potentially-different tables to produce a final assignment result; we first determine the state of each \textit{tracker} to be $match$ or $unmatch$ based on the table for the larger $c$, and then of the unmatched \textit{trackers}, ones having corresponding \textit{detections} in the table for the smaller $c$ are determined to be in the state of $potentially$-$match$. In this work, we use 1.0 and 0.1 for the larger $c$ and smaller $c$, respectively.

Figures~\ref{Fig_FxMatching}\textbf{b} and \ref{Fig_FxMatching}\textbf{f} illustrate two successive frames of a scene with one-to-one correspondence between \textit{trackers} and \textit{detections} ($N_{t} \!=\! N_{d}$) and a scene with an occlusion event where the object that has been tracked as \textit{tracker} ID=5 is occluded by (not detected due to) the object that has been tracked as \textit{tracker} ID=2 (${N_t} \!>\! {N_d}$). Regarding the scene \textbf{b}, Figs.~\ref{Fig_FxMatching}\textbf{c} and \ref{Fig_FxMatching}\textbf{d} show, respectively, the resultant assignment tables for the smaller $c$ and the larger $c$ in the format of arranging a bit configuration $\{b_{t,d}\}$ (a vector) in a $N_t \!\times\! N_d$ matrix. Figs.~\ref{Fig_FxMatching}\textbf{g} and \ref{Fig_FxMatching}\textbf{h} are the corresponding ones for the scene \textbf{f}. In the matrix, the sum of $b_{t,d}$ in a row (/a column) means the number of matched \textit{detections} (/\textit{trackers}) with the \textit{tracker} (/the \textit{detection}) corresponding to the row (/the column).

For the scene \textbf{b} (without occlusion), the assignment tables, \textbf{c} and \textbf{d}, for the smaller $c$ and the larger $c$ are the same, which satisfies the constraint for one-to-one correspondence (all the sums of $b_{t,d}$ in rows/columns are 1). For the scene \textbf{f} (with occlusion), the assignment table \textbf{h} for the larger $c$ satisfies the constraint, based on which \textit{tracker} ID=5 is first determined to be $unmatch$ by the arbiter (depicted in Fig.~\ref{Fig_FxMatching}\textbf{a}). Note that at this stage, there are two possibilities for the event that happened for the object being tracked by \textit{tracker} ID=5: frame-out or occlusion (crossing). The assignment table \textbf{g} for the smaller $c$ does not satisfy the constraint as both \textit{trackers} ID=5 and ID=2 are matched with \textit{detection} ID=4. Such a solution can be chosen by the Ising machine when the lowering in the objective function by having more matched \textit{trackers} compensates the increase in the penalty function by violating the constraint. Based on the table \textbf{g}, the arbiter finally determines the state of \textit{tracker} ID=5 to be $potentially$-$match$. Thus, the system detects the occurrence and its location of an occlusion event.

Figure~\ref{Fig_FxMatching}\textbf{e} shows the solution space of the decision variables (all the possible bit configurations) including constraint-violating/satisfying solutions. Assuming $N_{t} \!=\! N_{d}$ ($=N_{o}$) for simplicity, the size of the solution space is $2^{N_{o}^2}$, while the number of the possible solutions satisfying the constraint for one-to-one correspondence is $N_{o}!$, which is a small fraction of the solution space. The bit configurations, \textbf{c}, \textbf{d} and \textbf{h}, in Fig.~\ref{Fig_FxMatching} belong to the constraint-satisfying solutions, but the bit configuration \textbf{g} belongs to the constraint-violating solutions. The linear assignment algorithm (i.e., the Hungarian algorithm) searches for the solution within a subset consisting of constraint-satisfying solutions, while the Ising machine does within the whole space. Hence, we need the Ising machine to find the solution with considering feasible many-to-one assignments. It is essential to evaluate the degree of constraint violations by the penalty functions (the quadratic terms), Eqs.~\ref{Eq_pnlty1} and \ref{Eq_pnlty2}, instead of binary judgment by the equality constraints (Eqs.~\ref{Eq_const1} and \ref{Eq_const2}).

\subsection{Multiple object tracking}\label{sec_MOTsystem}

Many modern MOT systems~\cite{SORT,DeepSORT,JDE,FiarMOT,GIAOTracker,UniTrack,ByteTrack} rely on the Kalman filter framework~\cite{Kalman60} to estimate the dynamics of \textit{trackers}, where the statuses of \textit{trackers} are first predicted based on a motion model every frame and then corrected (updated) using the information of the matched \textit{detections} (a measurement result) at the frame (also see Fig.~\ref{Fig_MOTsystem} and Fig.~\ref{Fig_FxMatching}\textbf{a}). The framework enables more accurate motion estimation of \textit{trackers} by reflecting a series of measurements observed over time than those based on a single measurement that may include statistical noise and other inaccuracies.

Figures~\ref{Fig_MOTsystem} and \ref{Fig_FxMatching}\textbf{a} shows the block diagram of the MOT system consisting of camera, detector, predictor, corrector, associator, and assignor. The pseudocode (Algorithm~\ref{algo_part}) in the Methods section shows the procedure of information processing in the system. In this work, we adopt a MOT system called simple online and realtime tracking (SORT)~\cite{SORT} as a baseline since it is a representative one of modern MOT systems and simple to implement with vehicle-mountable computing resources, and then modify it with respect to two points. First, we change the assignment function in the assignor from the linear assignment assuming one-to-one correspondence (Hungarian algorithm) to the flexible assignment using an Ising machine and an arbiter (Fig.~\ref{Fig_FxMatching}\textbf{a}). Second, we modify the procedure in the corrector to utilize the newly introduced state of $potentially$-$match$ for tracking through occlusion.

The status data of $i$th \textit{tracker} include the data $T_{i}$:

\begin{equation} \label{Eq_age}
T_{i}=\lbrack \mathbf{r}_{i}, \dot{\mathbf{r}_{i}}, age_{i} \rbrack,
\end{equation}
where $\mathbf{r}_{i}$ represents the location and size of the bounding box for the \textit{tracker}, $\dot{\mathbf{r}_{i}}$ is the time derivative of $\mathbf{r}_{i}$, and $age_{i}$ is the age of the \textit{tracker} defined as the number of frames (the time) elapsed since it last acquired the matched \textit{detection}. At the begining of the procedure for each frame (see Algorithm~\ref{algo_part}), the statuses of \textit{trackers} at the frame are predicted by approximating the inter-frame displacements of objects and then $age$ for all \textit{trackers} are incremented by one. At the end of the procedure, the \textit{trackers} having $age$ greater than a predetermined lifetime, $max\_age$, are deleted. Processing for each \textit{tracker} in the corrector differs depending on the assignment result ($match$, $potentially\_match$, and $unmatch$) determined by the assignor based on the predicted \textit{trackers} and the \textit{detections} obtained at the frame. The matched \textit{tracker} is updated using the corresponding \textit{detection} (based on the Kalman filter theory), whose $age$ is set to be zero (refresh). The potentially-matched \textit{tracker} is not updated (the predicted status will be used as is in the next frame), but its $age$ is decreased by a predetermined constant, $anti\_aging$. The unmatched \textit{tracker} is not changed, which is deleted if its $age$ is greater than $max\_age$. In addition, a new \textit{tracker} is appended for each unmatched \textit{detection} (if exists). In this work, both $max\_age$ and $anti\_aging$ are set to be 5.

Once a \textit{tracker} is determined to be $potentially\_match$, it is not deleted at least for subsequent $anti\_aging$ frames. Furthermore, if it is determined to be $potentially\_match$ again during the period, its $age$ is further decreased by $anti\_aging$ ($age$ can be negative). Thus, it can find the corresponding \textit{detection} again after the occlusion event (an occlusion event usually lasts for several frames). We give those special treatments only for the specific \textit{trackers} (potentially-matched \textit{trackers}). In contrast, we can realize tracking through occlusion just by increasing $max\_age$ without introducing the state of $potentially\_match$. However, there are serious side effects. In this case, \textit{trackers} corresponding to objects that should be deleted (ex. frame-out objects or objects left behind buildings) remain for a long period, leading to unnecessary computing load for managing the increased \textit{trackers} in the Kalman filter framework and erroneous $match$ between those unfavorable \textit{trackers} and \textit{detections}.

\subsection{System architecture}\label{sec_System}

\begin{figure}[t]
\centering
\includegraphics[width=17.2 cm]{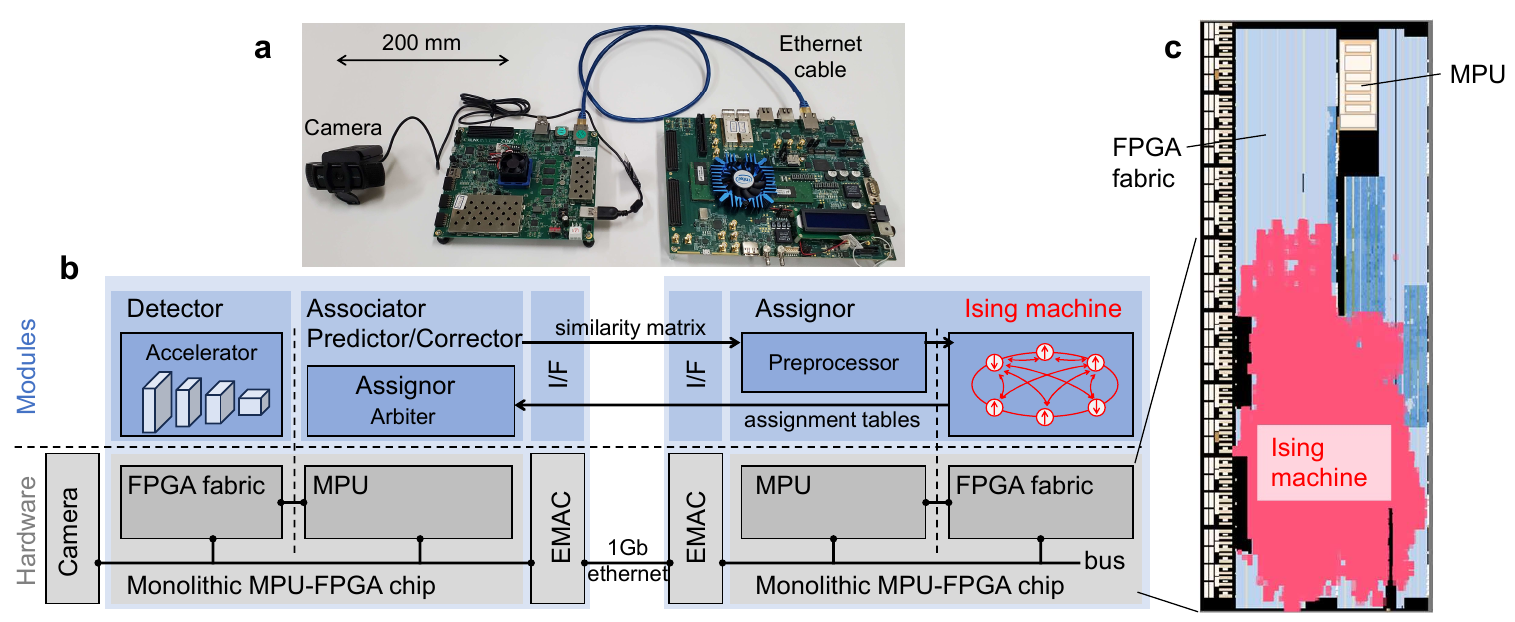}
\caption{Implementation of the MOT system with a vehicle-mountable computing platform. (\textbf{a}) A photograph of the system showing a camera and two computing boards with each having a monolithic MPU-FPGA chip. (\textbf{b}) Hardware configuration and the block diagram of modules implemented thereon. (\textbf{c}) Placement of the modules in the monolithic MPU-FPGA chip for the assignor. The custom circuit for simulated bifurcation as an embedded Ising machine is red highlighted in the FPGA fabric.}
\label{Fig_Implementation}
\end{figure}

To demonstrate the feasibility of the proposed MOT system under the constraints of size, power, and cost for in-vehicle computing platforms, we prototyped the MOT system with two vehicle-mountable computing boards with each having an embedded FPGA (as a parallel and programmable coprocessor) and a general-purpose micro processing unit (MPU). Figure~\ref{Fig_Implementation} shows the hardware configuration of the system and where the modules depicted in Fig.~\ref{Fig_MOTsystem} and Fig.~\ref{Fig_FxMatching}\textbf{a} are implemented. Of the modules, the computation-intensive ones that should be efficiently processed in terms of speed and power are the Ising machine and detector, which are hardwired (implemented as custom circuits) with the two FPGAs. The remaining modules are implemented as software objects which are processed with the MPUs.

The Ising machine of this work is 512-spin size with all-to-all spin-spin connectivity that allows us to set real numbers (32-bit precision) in any coupling coefficients ($J_{i,j}$). The QUBO problem (w/ the decision variable, $b_{i}\in{\{0,1\}}$) defined by Eqs.~\ref{Eq_def_b} and \ref{Eq_Cost} is represented as an Ising problem (w/ the decision variable, $s_{i}\in \{-1,1\})$ and then solved by the Ising machine. See the Methods section for the one-to-one correspondence between QUBO and Ising problems. The Ising machine supports up to 22 \textit{trackers} ($N_{t}$=22, $N_{d}$=22, $N_{t}N_{d}$=484). 

Simulated bifurcation (SB) is a highly-parallelizable metaheuristic algorithm for discrete optimization problem. For $N$-spin Ising problems with full connectivity, the maximum numbers of parallelizable operations in SB and simulated annealing (SA, a conventional metaheuristic)~\cite{SA83,isakov15} are, respectively, $N^2$ and $N$~\cite{FPL19,NatEle}. Custom-circuit implementations of SB~\cite{FPL19,NatEle,Kashimata24} with modern (island-style architecture~\cite{betz12} ) FPGAs have demonstrated a higher degree of computational parallelism than the problem size $N$. Of various variants of SB, we adopt the ballistic SB algorithm~\cite{sbm2} in this work, which is suitable for single-shot processing necessary for high-speed realtime systems~\cite{ISCAS20,ACCESS23a,ACCESS23b}. See the Methods section for the detail of simulated bifurcation. 

By using a scalable design of the accelerator for the ballistic SB (written in a high-level synthesis language) following similar circuit architectures in Refs.~\cite{FPL19,Hidaka23}, we built the embedded 512-spin SB-based Ising machine depicted in Fig.~\ref{Fig_Implementation}\textbf{c} with 2,048 parallel processing elements to compute 2,048 pair interactions, simultaneously in a clock cycle, included in the 512$\times$512 ones per SB time step (corresponding to the term of $\sum_{j}^{N}J_{i,j}x_j$ in Eq.~\ref{eq:y.bSB}). The computational parallelism was selected from a viewpoint of roughly estimated cost constraint for commercial vehicles [the number of logic elements $<$ 250K, the number of 32-bit digital signal processor units (DSPs) $<$ 400]. When $N_{step}$, a operational parameter for SB, is 400 (the case for this work), the time to obtain a solution with the SB-based Ising machine is 284 $\mu s$. The computation time to solve the QUBO problem twice per frame in this work (see Sec. \ref{sec_FxAssignment}) is 568 $\mu$s. The operating power of the Ising machine during realtime operation of the MOT system was measured to be 3.4 W. See the Methods section for the detail of the implementation of the MOT system.

\subsection{Demonstration}\label{sec_demo}

The MOT system with the embedded Ising machine, implemented with the vehicle-mountable computing boards, demonstrates a realtime processing speed and the enhanced functionality that needs NP-hard combinatorial optimization.

\begin{figure}[t]
\centering
\includegraphics[width=17.2 cm]{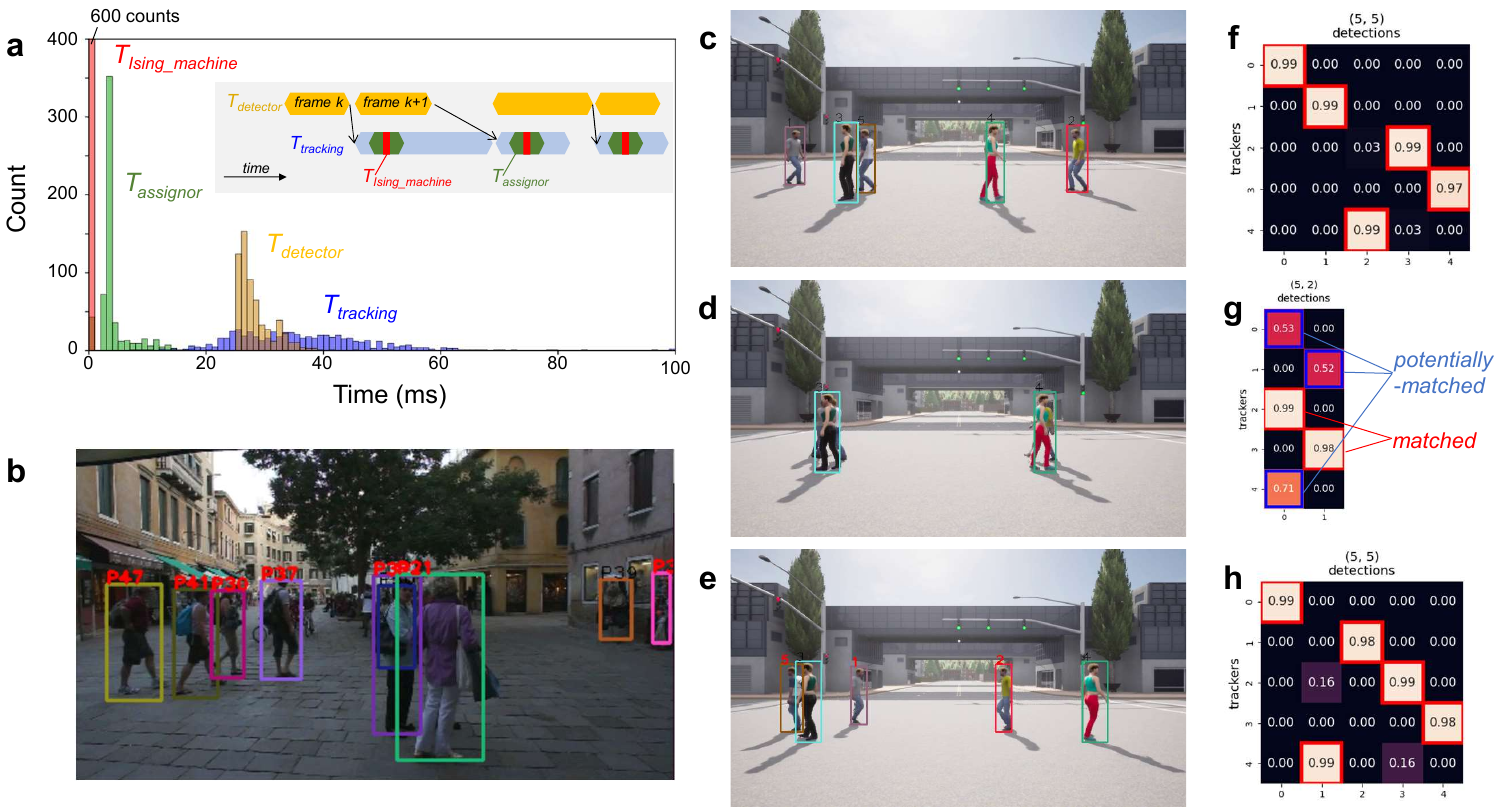}
\caption{Processing speed and functionality of the MOT system with the flexible assignment function. (\textbf{a}) Histogram of the calculation times of the modules in the MOT system when processing a MOT benchmark problem (a movie including 600 frames) named ``MOT17-02-FRCNN''~\cite{MOT17,MOT17data}. The inset illustrates the timing chart of the overlapped operation of the detector and the other tracking modules. (\textbf{b}) A scene extracted from ``MOT17-02-FRCNN'' with \textit{trackers} indicated as boxes. (\textbf{c, d, e}) Five-object tracking through a complex occlusion event (simultaneous occurrences of three-object crossing and two-object crossing). (\textbf{c, d, e}) shows, respectively, the frames \#50, \#80, and \#110 extracted from a movie (inc. 142 frames) provided as Supplementary Information 3. The boxes indicate \textit{trackers}. (\textbf{f, g, h}) Similarity matrixes with red and blue boxes meaning the assignment results between \textit{trackers} and \textit{detections}. (\textbf{f, g, h}) correspond to (\textbf{c, d, e}), respectively.}
\label{Fig_demo}
\end{figure}

Figure~\ref{Fig_demo}\textbf{a} depicts the histogram of the calculation times of the modules in the MOT system when processing a benchmark movie (inc. 600 frames) named ``MOT17-02-FRCNN''~\cite{MOT17,MOT17data}. Figure~\ref{Fig_demo}\textbf{b} illustrates the \textit{trackers} as the colored boxes for a scene extracted from the movie. The inset in Fig.~\ref{Fig_demo}\textbf{a} shows the timing chart of the MOT system. The operation of the detector ($T_{\mathrm{detector}}$) is overlapped with those of the other tracking modules ($T_{\mathrm{tracking}}$) including the assignor ($T_{\mathrm{assignor}}$). The processing time per frame is determined by $\max\{T_{\mathrm{detector}}, T_{\mathrm{tracking}}\}$, which is 44.2 ms on average corresponding to a processing speed of 23 frames per second. The calculation time of the embedded Ising machine ($T_{\mathrm{Ising\_machine}}$) is deterministic (568 $\mu$s) and minor compared to (or not affecting) the whole processing time. Note that $T_{\mathrm{assignor}}$ is measured by the MPU of the left board in Fig.~\ref{Fig_Implementation} and includes the times related to the inter-board communication, preprocessor and arbiter. The detector takes as input the video frames from either the onboard memory or the camera. See the Supplementary Information 2 for the realtime operation of the MOT system when using the camera.

Compared to the baseline (the original SORT~\cite{SORT}), the MOT system extended with the flexible assignment function shows a higher tracking performance in terms of a unified MOT evaluation metric called HOTA~\cite{HOTA}. The factor contributing to the enhanced HOTA score is an improvement in the association accuracy (not in the detection and location accuracies). See the Table~\ref{Tbl_Cmp} in the Methods section for details.

The proposed MOT system enables tracking through long-term and complex occlusion events which is not possible to be achieved with the baseline. We designed a movie scene having simultaneous occurrences of three-object crossing and two-object crossing. Here the three-object crossing is involved in a long-term occlusion event where two objects move in the same direction, and one overtakes another at small relative speed. The proposed system correctly tracks the five objects through those occlusion events, while the baseline fails to track. The movie which includes and compares the tracking results by the proposed system and the baseline is provided as the Supplementary Information 3. Figures~\ref{Fig_demo} \textbf{c}, \textbf{d}, and \textbf{e} shows, respectively, three scenes before, at, and after the complex occlusion event. During the simultaneous occurrences of three-object and two-object crossings, the states of \textit{trackers} ID=0, 1, 4 are $potentially$-$match$ as shown in Fig.~\ref{Fig_demo} \textbf{g}. The proposed system is capable of tracking through more complex occlusion events. The Supplementary Information 4 shows such an example having simultaneous occurrences in four locations of five-object crossings.

\section{Discussion}\label{sec_discussion}

We have demonstrated an in-vehicle multiple object tracking system with a flexible assignment function for tracking through long-term and complex occlusion events such as simultaneous multiple occurrences of many-object crossing. The flexible assignment, which is formulated as NP-hard discrete optimization, has been solved in real time by an SB-based Ising machine implemented with a vehicle-mountable computing platform (middle-range SoC-FPGA boards). The methodology of this work paves the way for enhancing the functionality of assignor beyond the linear assignment toward ones requiring computationally-hard combinatorial optimization.

There would be three possible directions of future works. First, the flexible assignment function could be combined with more advanced definitions of similarity~\cite{DeepSORT,JDE,FiarMOT,GIAOTracker,UniTrack,ByteTrack,TransMOT} in terms of utilizing some distances in feature space between \textit{trackers} and \textit{detections}. Second, the methodology of flexible assignment could be used for quadratic object function, $H_{\mathrm{object}} \!=\! - \sum_{(t, d)}\sum_{(t^{\prime},d^{\prime})}S_{t,d,t^{\prime},d^{\prime}}b_{t,d}b_{t^{\prime},d^{\prime}}$, in which $S_{t,d,t^{\prime}\neq t,d^{\prime}\neq d}$ represents a bonus/penalty for simultaneous matching of two pairs of ($t$, $d$) and ($t^{\prime}$, $d^{\prime}$) [corresponding to conditional likelihood or tradeoff relationship]. Note that in the QUBO formulation of this work, the penalty functions (Eqs.~\ref{Eq_pnlty1} and \ref{Eq_pnlty2}) are quadratic while the object function (Eq.~\ref{Eq_Obj}) is linear. Third, those in-vehicle computing units with embedded Ising machines could be applicable to various tasks other than MOT such as SLAM (simultaneous localization and mapping)~\cite{Mur-Artal17}, scheduling, or routing.

\clearpage

\section*{Methods}

\subsection*{MOT algorithm}

Algorithm~\ref{algo_part} shows the procedure of information processing in the MOT system with the flexible assignment, which consists of camera, detector, predictor, corrector, associator, and assignor illustrated in the block diagrams of Figures~\ref{Fig_MOTsystem} and \ref{Fig_FxMatching}\textbf{a}.

The detector detects objects belonging to the classes of ``car'' or ``person'' in the frame based on a realtime object detection algorithm called YOLO~\cite{YOLO16,YOLO17} to produce \textit{detections} (inc. the bounding boxes for detected objects). Similarly to the SORT~\cite{SORT}, the predictor is based on a linear constant velocity model and the associator uses intersection over union, $\mathrm{IOU}(t,d)$, of the bounding boxes for $t$th \textit{tracker} and $d$th \textit{detection} as the definition of similarity $S_{t,d}$. The assignor and corrector are detailed in the Main text.

\scriptsize
\begin{algorithm}[h]
\caption{MOT with the flexible assignment}\label{algo_part}
\For{$frame\ \mathbf{in}\ frame\_buffer$}{
$detections \leftarrow \mathrm{Detector}(frame)$ \\
$trackers \leftarrow \mathrm{Predictor}(trackers)$ \\
\For {$tracker\ \mathbf{in}\ trackers$}{
$tracker.age \leftarrow tracker.age + 1$\\
}
$similarity \leftarrow \mathrm{Associator}(trackers, detections)$\\
\tcp{\small Flexible assignment}
$assignment \leftarrow \mathrm{Assignor}(similarty)$\\
\tcp{\small Corrector}
\For{$tracker\ \mathbf{in}\ assignment.match$}{
$tracker \leftarrow \mathrm{Updater}(tracker, matched\_detection)$\\
$tracker.age \leftarrow 0$\\
} 
\For {$tracke\ \mathbf{in}\ assignment.potentially\_match$}{
$tracker.age \leftarrow tracker.age - anti\_aging$\\
}
\For {$detection\ \mathbf{in}\ assignment.unmatch$}{
$trackers.\mathrm{Append}(detection)$\\
}
\For {$tracke\ \mathbf{in}\ trackers$}{
\If {$tracker.age > max\_age$}{
$trackers.\mathrm{Delete}(tracker)$\\
}
}
}
\end{algorithm}
\normalsize

\subsection*{Ising and QUBO problems}

The $N$-variable Ising problem is to find a spin configuration that minimizes the Ising energy~\cite{barahona82}:

\begin{equation}\label{IsingEnergy}
H_{\mathrm{Ising}}=-\frac{1}{2}\sum_{i=1}^{N}\sum_{j=1}^{N}J_{i,j}s_{i}s_{j}+\sum_{i=1}^{N}h_{i}s_{i},
\end{equation}
where $s_{i}\;(\in \{-1,1\})$ is the $i$th Ising spin, $J_{ij}$ is the coupling coefficient between the $i$th and $j$th spins, and $h_{i}$ is the bias coefficient (local field) for the $i$th spin.

The QUBO problem ($b_{i}\in{\{0,1\}}$) with the cost function,

\begin{equation}
H_{\mathrm{QUBO}}=\sum_{i}^{N}\sum_{j}^{N}Q_{i,j}b_{i}b_{j},
\end{equation}
is represented as the Ising problem using following conversions.
\begin{align}
s_{i}&=2b_{i}-1,
\label{eq:conv1}\\
J_{i,j}&=
\begin{cases}
-\frac{Q_{i,j}}{2} & (\text{if}\;i\neq j),\\
0 & (\text{if}\;i= j),
\end{cases}
\label{eq:conv2}\\
h_{i}&=\sum_{j}^{N}\frac{Q_{i,j}}{2}.
\label{eq:conv3}
\end{align}

\subsection*{Simulated bifurcation}

Simulated bifurcation (SB)~\cite{sbm1,sbm2} is a quantum-inspired~\cite{sbm1,Qinspired24}, highly-parallelizable~\cite{FPL19,NatEle,Kashimata24}, metaheuristic algorithm for computationally-hard combinatorial (or discrete) optimization. SB-based Ising machine belongs to a group of oscillator-based Ising machines~\cite{honjo21,kalinin20,PoorCIM19,moy22,albertsson21,wang21,SimCIM21,Graber24}. The SB finds the optimal (exact) or near-optimal solution of the Ising problem by simulating the time-evolution process of coupled nonlinear oscillators according to the Hamilton's equations of motion (without energy-dissipative or noise-based mechanisms). The SB has several variants including adiabatic SB, ballistic SB, and discrete SB, which differ in terms of nonlinearity~\cite{Nonlinearity21} and discreteness~\cite{sbm2}. 

In the SB, the $i$th nonlinear oscillator corresponds to the $i$th Ising spin and its state is described by the position and momentum ($x_i$, $y_i$). The update procedure of $x_i$ and $y_i$ for the ballistic SB, used in this work, is as follows~\cite{sbm2}.

\begin{align}
y_i^{t_{k\!+\!1}} &\gets y_i^{t_k} + [-(a_0-a^{t_k})x_i^{t_k} -\eta h_i + c_0\sum_{j}^{N}J_{i,j}x_j^{t_k}]\Delta_t,
\label{eq:y.bSB}\\
x_i^{t_{k\!+\!1}} &\gets x_i^{t_k} + a_0 y_i^{t_{k\!+\!1}}\Delta_t,
\label{eq:x.bSB}\\
(x_{i}^{t_{k\!+\!1}}, y_{i}^{t_{k\!+\!1}})&\gets
\left\{ 
\begin{alignedat}{2} 
(\mathrm{sgn}(x_{i}^{t_{k\!+\!1}}), 0) & \;\;(\text{if}\;|x_{i}^{t_{k\!+\!1}}|>1), \\
(x_{i}^{t_{k\!+\!1}}, y_{i}^{t_{k\!+\!1}}) & \;\;(\text{if}\;|x_{i}^{t_{k\!+\!1}}|\le1),
\end{alignedat} 
\right.
\label{eq:wall}
\end{align}
where $a_0$, $c_0$ and $\eta$ are positive constants, $a^{t_k}$ is a control parameter increasing from zero to $a_0$, and $\mathrm{sgn}(x) (=\pm 1)$ is the sign function. Eq.~\ref{eq:wall} is a nonlinear transfer function~\cite{Nonlinearity21}, physically corresponding to a perfectly inelastic wall existing at $x=\pm 1$. The time increment is denoted as $\Delta_t$, and thus, $t_{k+1}=t_k+\Delta_t$. After iterating the update procedure for the predetermined time steps ($N_{\rm step}$), the $i$th position $x_{i}$ is digitized to be the $i$th spin ($\pm 1$) by taking the sign of $x_{i}$. In this work, $a_0$=1, $c_0=\eta=0.8$, $\Delta_t =0.3$, and $N_{\rm step}=400$.

The ballistic SB allows us to obtain better-quality solutions faster than the simulated annealing (SA) algorithm for academic benchmark problems~\cite{sbm2} and practical problems~\cite{Matsumoto22,ISCAS20,ACCESS23a,ACCESS23b,Hidaka23}.

\subsection*{Implementation}

For implementing the proposed MOT system, we used two vehicle-mountable SoC (System-on-Chip)-FPGA boards with each having a monolithic MPU-FPGA chip. 

One is the Intel Arria10 SX SoC Development Kit (DK-SOC-10AS066S-D) with an 10AS066N3F40E2SG1 monolithic chip including a dual-core ARM Cortex-A9 MPCore processor and an embedded FPGA, where the embedded FPGA (660K logic elements, 4-input LUT equivalent) has 251,680 adaptive logic modules (ALMs) including 251,680 adaptive look-up-tables (ALUTs, 6-input LUT equivalent) and 1,006,720 flip-flop registers, 2,131 20Kbit-size RAM blocks (BRAMs)~\cite{BRAM}, and 3,374 18-bit$\times$19-bit multiplier (DSPs). 

The SB-based Ising machine was coded in a high-level synthesis (HLS) language (Intel FPGA SDK for OpenCL, ver. 18.1) and implemented with the embedded FPGA. Table \ref{Tbl_Imp} summarizes the details of the architecture and the implementation result. The system clock frequency ($F_\mathrm{sys}$) determined as a result of circuit synthesis, placement, and routing is 254 MHz. The operating power of the Ising machine is 3.4 W, which was observed by the PowerMonitor tool that uses the MAX V CPLD on the board to measure currents running on the power rails to the target FPGA. Software objects processed with the MPU were written in C/C++ programming language and executed on Linux OS (Angstrom v2014.12).

The other board is the AMD Zynq UltraScale+ MPSoC (ZCU104) with a XCZU7EV-2FFVC1156 monolithic chip including a quad-core ARM Cortex-A53 MPCore processor and an embedded FPGA, where the embedded FPGA (504K logic cells, 4-input LUT equivalent) has 28,800 configurable logic block (CLB) including 230,400 adaptive look-up-tables (ALUTs, 6-input LUT equivalent) and 460,800 flip-flop registers, 312 36Kbit-size RAM blocks (BRAMs)/96 288Kbit-size RAM blocks (UltraRAMs), and 1,728 27-bit$\times$18-bit multiplier (DSPs). 

The detector implemented with the embedded FPGA is a custom circuit for YOLOv2~\cite{YOLO17}, yolov2\_voc\_pruned\_0\_77, provided by Advanced Micro Devices, Inc. Software objects processed with the MPU were written in Python (ver. 3.8) and executed on Linux OS (Ubuntu Desktop 20.04.5/6 LTS).

The two boards are equipped with ethernet media access controllers (EMACs) and connected with a 1 Gb ethernet cable (in the UPD protocol).

\scriptsize
\begin{table}
\caption{Implementation of the SB-based Ising machine}
\label{Tbl_Imp}
\centering
\begin{tabular}{l|c}
\toprule
\textbf{Architecture}&\\
machine size (\# of spins)&512\\
connectivity (spin-spin coupling)&full connection\\
\# of parallel processing elements (PEs) &2,048\\
\midrule
\textbf{Resource usage}&\\
ALM (count) [(\%)] &81,341\,/\,251,680 [32\%]\\
(Logic element [(\%)]) & (211K\,/\,660K [32\%])\\
BRAM (bit) [(\%)] &2,377,896\,/\,43,642,880 [5\%]\\
32-bit DSP (count) [(\%)] &325\,/\,1,687 [19\%]\\
\midrule
\textbf{Speed features}&\\
$F_\mathrm{sys}$ (MHz)&254\\
$T_\mathrm{step}$ ($\mu$s per SB time step)&0.71\\
\toprule
\end{tabular}
\end{table}
\normalsize

\subsection*{Evaluation}

We compared the proposed MOT and the SORT with a common parameter of $max\_age$ (=5) in terms of a MOT evaluation metric named HOTA (higher order tracking accuracy)~\cite{HOTA} for seven MOT benchmark problems (movies), MOT17-\{02, 04, 05, 09, 10, 11, 13\}-FRCNN~\cite{MOT17,MOT17data}. HOTA is a unified and balanced metric consisting of three sub-metrics: AssA (association accuracy), DetA (detection accuracy), and LocA (localization accuracy). Table \ref{Tbl_Cmp} summarizes the measured results for the proposed MOT, where the numbers (percentages) in parentheses show the ratios versus the SORT. The overall HOTA score is enhanced for the proposed MOT owing to the improvements in AssA by the flexible assignment function.

\scriptsize
\begin{table}
\caption{Comparison between the proposed MOT and the SORT}
\label{Tbl_Cmp}
\centering
\begin{tabular}{l|c|ccc}
\toprule
Benchmark &\textbf{HOTA}$\uparrow$&AssA$\uparrow$&DetA$\uparrow$&LocA$\uparrow$\\
\midrule
MOT17-02&\textbf{33.4 (106\%)}&37.1 (110\%)&30.5 (103\%)&86.6 (98\%)\\
MOT17-04&\textbf{51.8 (103\%)}&56.0 (108\%)&48.1 (99\%)&89.6 (99\%)\\
MOT17-05&\textbf{42.4 (99\%)}&47.2 (99\%)&38.1 (98\%)&84.5 (99\%)\\
MOT17-09&\textbf{49.5 (113\%)}&49.4 (125\%)&49.7 (101\%)&90.6 (99\%)\\
MOT17-10&\textbf{34.6 (89\%)}&28.9 (88\%)&42.1 (91\%)&84.4 (98\%)\\
MOT17-11&\textbf{52.5 (106\%)}&54.6 (115\%)&50.5 (97\%)&91.0 (99\%)\\
MOT17-13&\textbf{41.1 (97\%)}&42.2 (98\%)&40.6 (97\%)&84.0 (99\%)\\
\midrule
Overall&\textbf{45.6(102\%)}&48.5 (106\%)&43.3 (98\%)&88.0 (99\%)\\
\toprule
\end{tabular}
\end{table}
\normalsize

\section*{Data availability}

The authors declare that all relevant data are included in the manuscript. Additional data are available from the corresponding author upon reasonable request.


\section*{Acknowledgments}

The authors would like to thank Yoshihiko Isobe, Masataka Hirai, Ryo Hidaka, Yutaka Yamada, Yutaro Ishigaki, Tomoya Kashimata, Kei Nihei, Ryota Umino, Hayato Goto, Hiroomi Chono, Akiko Yuzawa, Sakie Nagakubo for the fruitful discussion and their support.

\section*{Competing interests}

K.T., Y.H., and M.Y. are included in inventors on two U.S. patent applications related to this work filed by the Toshiba Corporation (no. 17/249353, filed 20 February 2020; no. 18/456494, filed 27 August 2023). The authors declare that they have no other competing interests.

\section*{Author contributions}

All the authors contributed to the whole aspects of this work, with each making the following major contribution. K.T., K.O., and H.F. conceived and managed the project. K.T. devised the flexible assignment method and wrote the manuscript. Y.H. architected the whole system and designed the custom circuit of SB-based Ising machine. M.Y. implemented and integrated the system. K.O. considered the driving scenario. Y.H. and K.O. evaluated the system.

\section*{Additional information}

\subsection*{Supplementary information}

The online version contains supplementary materials.
\begin{itemize}
\item Supplementary information 1: A document that explains the Supplementary information 1 to 4. 
\item Supplementary information 2: A movie that demonstrates the realtime operation of the proposed MOT sytem.
\item Supplementary information 3: A movie of the five-object tracking corresponding to Figs.~\ref{Fig_demo}(\textbf{c, d, e}).
\item Supplementary information 4: A movie that demonstrates tracking through a complex occlusion event (simultaneous occurrences in four locations of five-object crossings).
\end{itemize}
\textbf{Note that the preprint (arXiv) version does not contain the Supplementary information 2, 3, and 4.}

\clearpage
\section*{Supplementary Information 1}

\subsection*{Supplementary information 1}
This document, which explains the Supplementary information 1 to 4.

\subsection*{Supplementary information 2}
A movie (20 seconds) that demonstrates the realtime operation of the proposed MOT sytem.

\begin{figure}[h]
\centering
\includegraphics[width=8.6 cm]{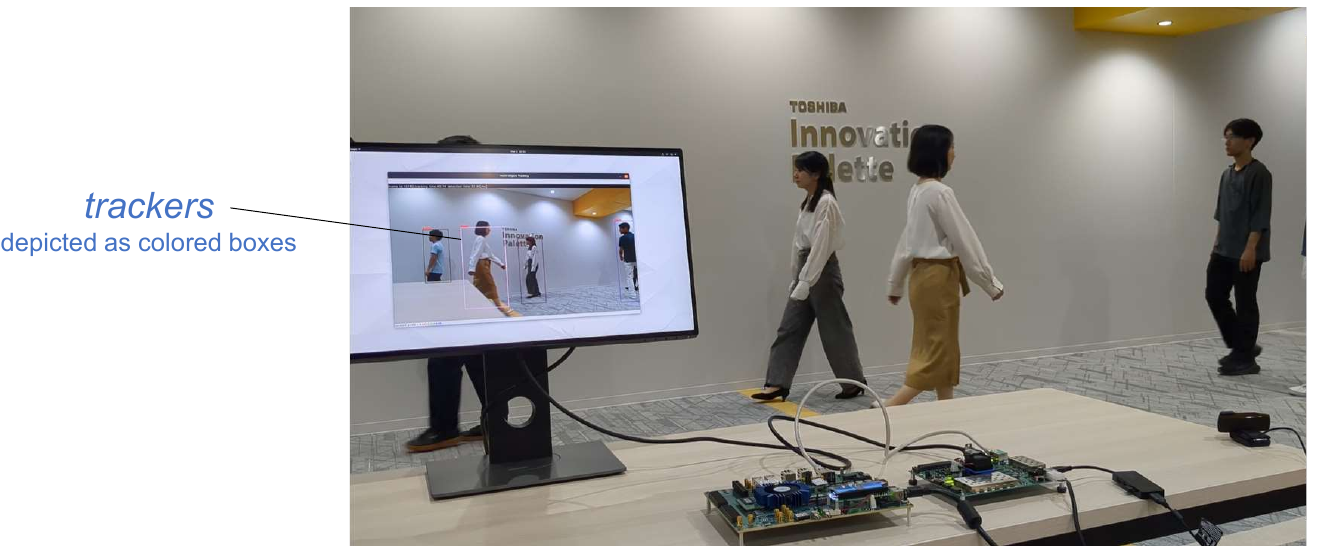}
\caption{A scene extracted from the Supplementary information 2}
\label{Fig_Supp2}
\end{figure}

\subsection*{Supplementary information 3}
A movie (14 seconds) of the five-object tracking corresponding to Figs. 4(\textbf{c, d, e}) in the Main text.

\begin{figure}[h]
\centering
\includegraphics[width=8.6 cm]{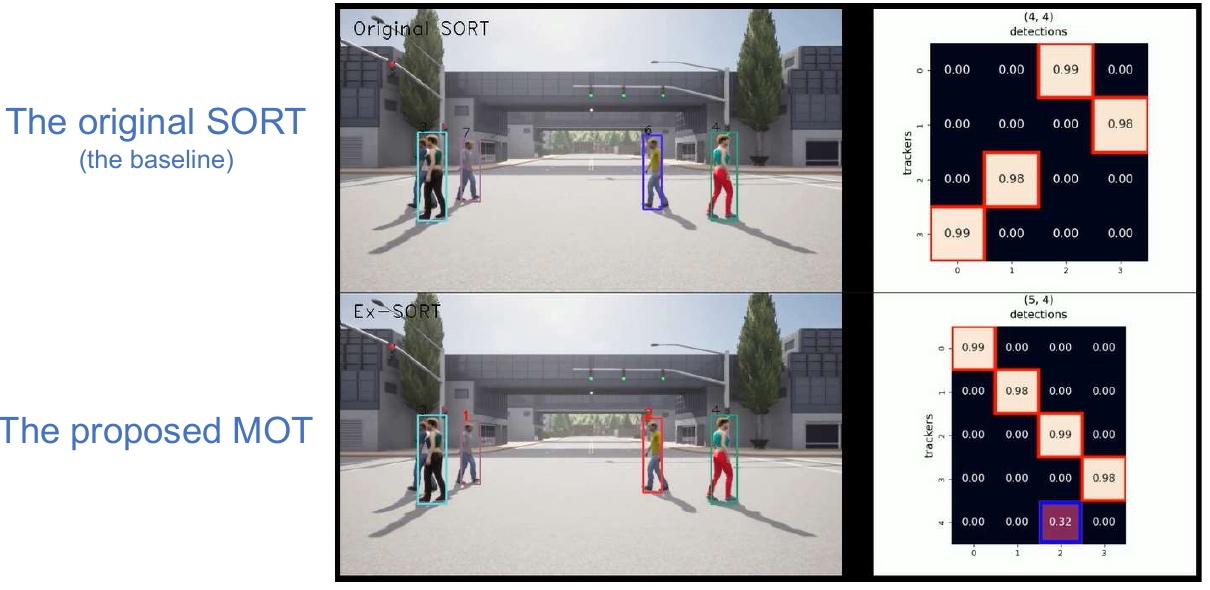}
\caption{A scene extracted from the Supplementary information 3}
\label{Fig_Supp3}
\end{figure}

\subsection*{Supplementary information 4}
A movie (19 seconds) that demonstrates tracking through a complex occlusion event (simultaneous occurrences in four locations of five-object crossings).

\begin{figure}[h]
\centering
\includegraphics[width=8.6 cm]{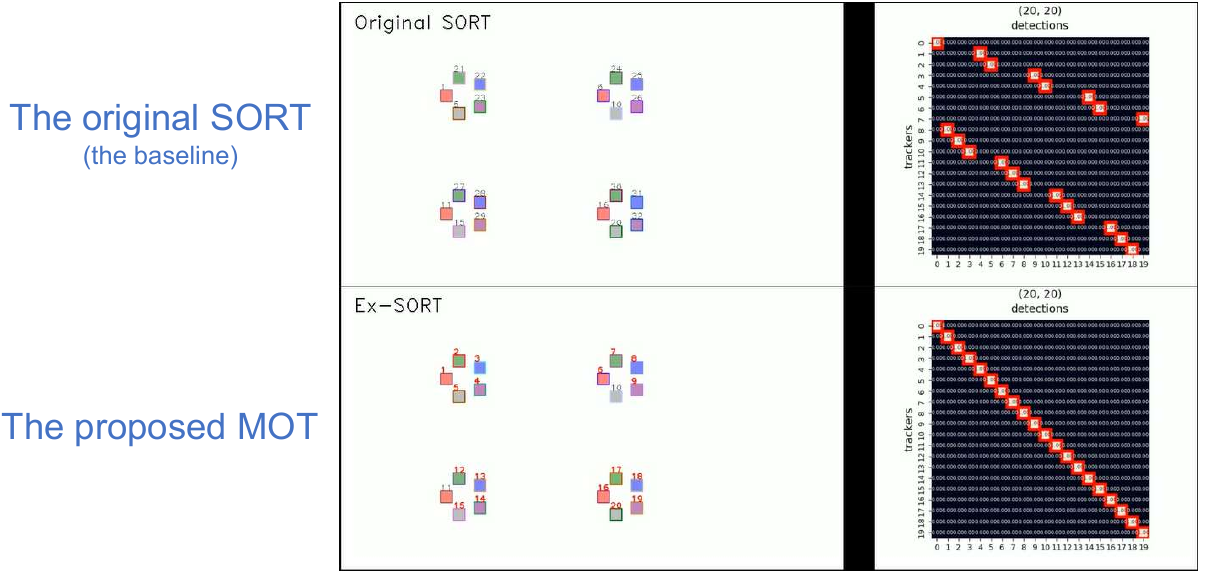}
\caption{A scene extracted from the Supplementary information 4}
\label{Fig_Supp4}
\end{figure}


\end{document}